\documentclass[journal]{IEEEtran}

\usepackage{booktabs}
\usepackage{placeins}
\usepackage{color,soul}

\usepackage{float}
\ifCLASSINFOpdf
\usepackage[pdftex]{graphicx}
\else
\fi
\usepackage{amsmath}
\usepackage{bm}
\begin{document}
%
% paper title
% Titles are generally capitalized except for words such as a, an, and, as,
% at, but, by, for, in, nor, of, on, or, the, to and up, which are usually
% not capitalized unless they are the first or last word of the title.
% Linebreaks \\ can be used within to get better formatting as desired.
% Do not put math or special symbols in the title.
\title{TAPER: Time-Aware Patient EHR  Representation}
%
%
% author names and IEEE memberships
% note positions of commas and nonbreaking spaces ( ~ ) LaTeX will not break
% a structure at a ~ so this keeps an author's name from being broken across
% two lines.
% use \thanks{} to gain access to the first footnote area
% a separate \thanks must be used for each paragraph as LaTeX2e's \thanks
% was not built to handle multiple paragraphs
%

\author{Sajad~Darabi,~\IEEEmembership{Student Member,~IEEE,}
        Mohammad~Kachuee,~\IEEEmembership{Student Member,~IEEE,}
        Shayan~Fazeli,~\IEEEmembership{Student Member,~IEEE,}
        and~Majid~Sarrafzadeh,~\IEEEmembership{Fellow,~IEEE}% <-this % stops a space
\thanks{Authors are with the Computer Science Department, University of California Los Angeles, CA, 90025}% <-this % stops a space
% \thanks{J. Doe and J. Doe are with Anonymous University.}% <-this % stops a space
% \thanks{Manuscript received April 19, 2005; revised August 26, 2015.}

}

\maketitle

% As a general rule, do not put math, special symbols or citations
% in the abstract or keywords.
\begin{abstract}
Effective representation learning of electronic health records is a challenging task and is becoming more important as the availability of such data is becoming pervasive.  The data contained in these records are irregular and contain multiple modalities such as notes, and medical codes. They are preempted by medical conditions the patient may have, and are typically recorded by medical staff. Accompanying codes are notes containing valuable information about patients beyond the structured information contained in electronic health records. We use transformer networks and the recently proposed BERT language model to embed these data streams into a unified vector representation.
% Further, as a patient may visit the clinic multiple times for differing reasons, the data entered is accompanied by timestamps, in which, their current situation may have been compounded by earlier diagnosis in earlier visits presenting challenges for modeling. In this work, we propose encoding scheme for embedding patients taking into account time, code, and textual information. 
The presented approach effectively encodes a patient's visit data into a single distributed representation, which can be used for downstream tasks. Our model demonstrates superior performance and generalization on mortality, readmission and length of stay tasks using the publicly available MIMIC-III ICU dataset. Code avaialble at https://github.com/sajaddarabi/TAPER-EHR
\end{abstract}

% Note that keywords are not normally used for peerreview papers.

% For peer review papers, you can put extra information on the cover
% page as needed:
% \ifCLASSOPTIONpeerreview
% \begin{center} \bfseries EDICS Category: 3-BBND \end{center}
% \fi
%
% For peerreview papers, this IEEEtran command inserts a page break and
% creates the second title. It will be ignored for other modes.
\IEEEpeerreviewmaketitle

\section{Introduction}
\label{sec:introduction}
% The very first letter is a 2 line initial drop letter followed
% by the rest of the first word in caps.
% 
% form to use if the first word consists of a single letter:
% \IEEEPARstart{A}{demo} file is ....
% 
% form to use if you need the single drop letter followed by
% normal text (unknown if ever used by the IEEE):
% \IEEEPARstart{A}{}demo file is ....
% 
% Some journals put the first two words in caps:
% \IEEEPARstart{T}{his demo} file is ....
% 
% Here we have the typical use of a "T" for an initial drop letter
% and "HIS" in caps to complete the first word.

% introduction 2 page
\IEEEPARstart{E}{lectronic} health records (EHR) are commonly adopted in hospitals to improve patient care. In an intensive care unit (ICU), various data sources are collected on a daily basis as preempted by medical staff as the patient undergoes care in the unit. The collected data consists of data from different modalities: medical codes such as diagnosis which are standardized by well-organized ontology's like the International Classification of Disease (ICD)\footnote{http://www.who.int/classification/icd/en} and medication codes standardized using National Drug Codes (NDC)\footnote{http://www.fda.gov}. Similarly, at various stages of the patient's care physicians input text noting relevant events to the patient's prognosis. Additionally, lab tests and bedside monitoring devices are used to collect signals each of which are collected at varying frequencies for a quantitative measure of the patient care. There is a wealth of information contained within EHRs that has a significant potential to be used to improve care.  Examples of inference tasks using such data include estimating the length of stay, mortality, and readmission of patients \cite{teres1987validation, campbell2008predicting}. 

\begin{figure}
\centering
\includegraphics[scale=1.2]{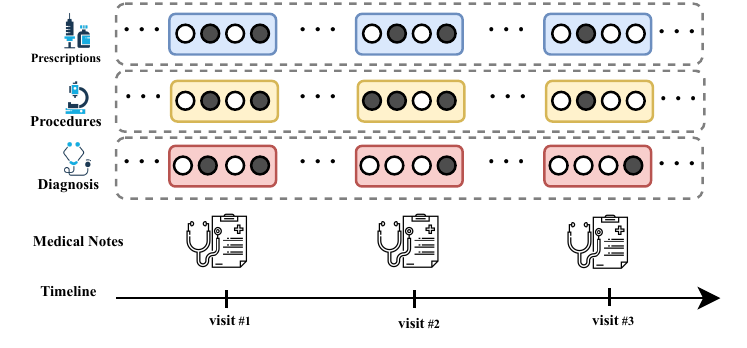}
\caption{Patient timeline during an ICU visit where different data points are collected. These include prescriptions, diagnosis codes, procedure codes and medical notes.}
\label{fig:my_label}
\end{figure}

The traditional approach for healthcare analysis has mainly focused on classical methods for extracting hand-engineered features and designing rule-based systems. More recently, deep learning has demonstrated state of the art results on a varying set of tasks, in which learning intermediate representation is at the heart of all these analysis \cite{bengio2013representation}. This representation can be obtained without domain-specific expertise by leveraging available EHR data. Although such methods have demonstrated great performance on image, audio datasets, leveraging deep learning techniques on healthcare data present new challenges as the data entered are sparse and contain different modalities. 

As is common in natural language processing tasks, the typical method for embedding medical codes and text could be through the use of one-hot vectors; though these are naturally high-dimensional and sparse resulting in poor performance. To alleviate this, the idea of learning a distributed representations as applied to natural language processing \cite{mikolov2013} has been also applied on medical data \cite{DBLP:journals/corr/abs-1806-02873}. Such methods share a common intuition that similar medical codes should share a similar context. Additionally, codes have varying temporal context, as such patients may have multiple visits with a similar set of codes. As an example, flu is short-lived whereas a diagnosis code for a more terminal disease such as cancer has a longer scope and hence will be present on all of the patient's visits. Due to the varying temporal context, it is also important to take into account the temporal scope of codes and texts assigned \cite{cai2018medical}. This demands for a model which takes the sequential dependencies of the patient's visits into account. 

To capture the sequential dependencies present in medical data recurrent neural networks (RNNs), and Long Short-Term Memory (LSTM) have become the go-to model. RNN auto-encoder models are commonly augmented with attention mechanisms allowing the model to attend to specific time steps either through soft/hard attentions resulting in improvement and interpretability in the final representation obtained  \cite{bahdanau2014neural, luong2015effective}. In NLP tasks such attention mechanisms are not required to be causal in time and hence can attend to both past representation as well as future representations to generate the current representation. However, in a healthcare setting, it is desirable to have the representation be causal in time as clinical decisions are made sequentially. Recently, transformer models \cite{vaswani2017attention} were proposed for natural language processing tasks, and have shown impressive results. It uses self-attention and as the model creates intermediate representations of the input it attends to its representation at previous and future timesteps when considering the present representation.

%% earlier work focus on codes or text few on both.. text is informative, medical codes informative.. combine into a single representation
Majority of patient representation work has solely focused on embedding medical codes or text as a patient representation for downstream tasks but not both. To address this, we study the use of transformer networks to embed structured medical code data as well as a language model to embed the text portion of visits. In this work, we propose to combine the medical representation from text and medical codes into a unified representation which can then be used for downstream prediction tasks. Lastly, the presented study takes into account the temporal context of a visit and embeds subsequent visits given the patient's history. In the following sections, we briefly go over related works and present our method followed by experimental setup and results.  

We have made our code and preprocessing steps available\footnote{https://github.com/sajaddarabi/TAPER-EHR}.

% You must have at least 2 lines in the paragraph with the drop letter
% (should never be an issue)

\section{Related Works}
\label{sec:related_works}

The idea of learning embeddings for sparse one-hot vector data types using back-propagation was presented in \cite{rumelhart1988learning}. Follow up work learned these embedding using neural networks \textit{Bengio et al.}\cite{bengio2003neural}. Since its success in NLP \textit{Mikolov et al.}\cite{mikolov2013} for language modeling, similar approaches have been used in the health domain. The two intuitive methods that are commonly used for word embedding are (1) skip-gram where a current word is said to predict surrounding words, (2) continuous bag of words (CBOW) where a set of words are made to predict a center word. In \textit{Choi et al. } \cite{choi2016multi} medical codes (diagnosis, procedure, and medications) are concatenated as one-hot vectors and embedded using the skip-gram model. The intuition behind skip-gram model is: codes in a visit should be predictive of its surrounding immediate patient visit codes as well. They also present an additional code loss term as regularization to the objective. It follows the intuition that codes in a visit should also predict one another. Similar to \cite{choi2016multi}, \textit{Nguyen et al.} \cite{nguyen2016deepr} use the concatenation of code representation and apply 1D convolutional network to obtain a visit representation. Follow up work augment these methods with external ontology's and attention on such external data sources when learning the representation \cite{Choi:2017:GGA:3097983.3098126, zhang2018patient2vec}. 

More recent work takes advantage of the hierarchical structure present in medical codes as they are assigned to a patient. For example in \cite{MIME2018}, the final visit representation is created hierarchically: first codes are embedded at a treatment level where a set of medication/procedures codes predict diagnosis codes, followed by diagnosis level where the representations at this level are made to predict next visits codes. Empirically the method can learn from a small set of samples and outperform earlier methods as presented on their proprietary datasets. Although these methods achieve reasonable results they do not explicitly model temporal context. This is important as certain clinical codes are short-lived whereas others could be long-lived or be permanent. As a result, certain codes should not be regarded as the context of one another although they may occur in the same visit or subsequent visit. This has been studied in \textit{Cai et al.}\cite{cai2018medical} where they train a CBOW model with temporal attention on the code representations. Most of these methods are concerned with medical code embedding and disregard physician notes which can potentially attribute to an improved representation.

Medical notes contain a vast amount of information but have not been studied adequately, especially for downstream tasks in the medical setting. Most works have focused on clinical concept embedding instead. In a clinical setting, nurses and doctors document patients progress. As notes are typically recorded using medical jargon, they do not necessarily follow the common grammatical structure found in English text. As such building, a representation from medical text using hand-engineered features is a challenge \cite{galvan-etal-2018-investigating}. For example in \cite{rajkomar2018scalable} the authors use unstructured EHR data and learn semi-supervised patient representation which are then evaluated on downstream tasks such readmission, mortality and length of stay. Along this line of work, in \cite{textrepresentation} text from EHR are used to embed patient text by predicting billing codes and averaged for downstream tasks using neural networks. Similarly, in \cite{liu2018deep}, the authors evaluate different models for embedding clinical notes such as CNNs, LSTMs and evaluate them on chronic disease prediction.  Although they showed good results their models are not expressive enough to capture all of the salience present in clinical text. To this end, a recent model namely Bidirectional Encoder Representation Transformers (BERT) language model presented by \textit{Devlin et al.}\cite{devlin2018bert} have recently outperformed previous methods on many benchmarks. This model was used in \cite{clinicalbert} to embed medical notes of patients. They evaluate their representations predictive performance on downstream tasks showing state of the art results compared to other methods.  

Few works have studied the combination of both text and clinical codes. Previous work \cite{bai2018ehr}, trains skip-gram and word2vec models to jointly embed clinical concepts and clinical text into a unified vector. Similarly, in \cite{mullenbach2018explainable} they use clinical text to predict clinical concepts. Although both text and code are taken into account, they are different from transferring the joint representation to downstream tasks that is the focus of our work.

% Method 3 pages

\section{Method}
\label{sec:method}
%figure for overview of method
\begin{figure}
\centering
\includegraphics[scale=1.35]{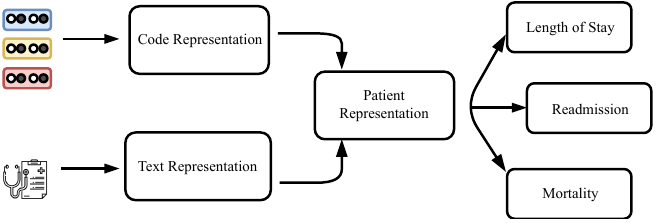}
\caption{Overview of method used to obtain patient visit representation.}
\label{fig:overview}
\end{figure}
The objective of this work is to create a distributed representation for a patient based on text and medical codes. This representation is then fed to classifier for predictive analytics tasks such as mortality, length of stay, and readmission (Fig.~\ref{fig:overview}). We split the training into two steps, (1) Skip-gram model using transformer networks to learn medical code representation, (2) a BERT model is trained on medical notes and the resulting representations at a time step are summarized using auto-encoder architectures \cite{DBLP:journals/corr/abs-1810-04805, vincent2008extracting, mkdenoising}. The final representation for a patient is a concatenation of these two. We discuss the approach in more detail in the following subsections.

To present the problem setting, the sequence of EHR data under consideration consists of a finite set of medical concepts $\mathcal{C} = \mathcal{M} \cup \mathcal{D} \cup \mathcal{P}$, where $\mathcal{M}$ is the set of medication codes, $\mathcal{D}$ is the set of diagnosis codes, and $\mathcal{P}$ is the set of procedure codes. Accompanying the codes are medical notes $\mathcal{T}$. We denote a patients longitudinal data as $\mathcal{D}^T= \{ (c_0, t_0), \cdots (c_T, t_T)\}$ with $T$ visits where $c_i$ and $t_i$ correspond to the codes and texts assigned respectively within the same visit window.

\subsection{Medical Code Embedding}
% figure for medical code embedding
\begin{figure}
\centering
\includegraphics[scale=1.8]{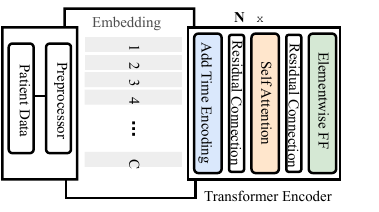}
\caption{The code representation module is a transformer encoder, which takes as input patient clinical codes. The embedding matrix is a $\Re^{d\times C}$ matrix. Clinical codes are embeeded using the embedding matrix, which is then passed to the transformer encoder block.}
\label{fig:code_encoder}
\end{figure}
We use skip-gram model to learn code embeddings as it is able to capture relationships and co-occurence between codes. We briefly review skip-gram model presented by \textit{Mikolov et al.}\cite{mikolov2013}. Given a sequence of codes $\{c_1, c_2, \dotsc, c_T\}$, where each code vector is a binary vector $c_t \in \{0,1\}^{|\mathcal{C}|}$, the model is tasked to predict the neighbouring codes given a code $c_t$. The objective can be written as 

\begin{align}
\frac{1}{T}\sum_{t=1}^T\sum_{-w\leq j\leq w, j\neq0} c_{t+j}^\top \text{log}(\hat{c_t}) + (1 - c_{t+j}^\top)\text{log}(1 - \hat{c}_t)
\end{align}

Here $w$ is the context window, and the softmax function is used to model the distribution $p(c_{t+j}|c_t)$. We use multiple transformer encoder layers with self-attention mechanism as the model for skip-gram. This model is then trained on medical code sequence $S = \{c_0, \cdots c_T\}$ by stacking code vectors into a matrix $K \in \{0,1\}^{T \times |\mathcal{C}|}$. The resulting set of codes are then converted into a set of embedding codes $e_{c_t} \in \Re^d$ using an embedding matrix $W \in \Re^{|\mathcal{C}| \times d}$. The embedding for the set of codes $c_t$ at visit $t$ is obtained as

\begin{align}
e_{c_t} = W^Tc_t
\end{align}

As the model does not contain any recurrence or convolution, to enable the model to make use of the ordering we need to inject information about the relative positioning of each embedding. This is done by adding to each embedding position a sinusoidal with frequency as a function of its timestamp $t$ as suggested by the original transformer network. This signal acts as positional-dependent information which the model could use to incorporate time. The model is summarized in Fig.~\ref{fig:code_encoder}. We stack multiple transformer layers following on top of the embedding matrix. By transformer layer, we mean a block containing the multi-head self-attention sub-layer followed by feed-forward and residual connections. For more details on this refer to \cite{vaswani2017attention} and the \textit{tensor2tensor} library\footnote{https://github.com/tensorflow/tensor2tensor}.  As multi-head attention can attend to future time steps, to ensure that the model's predictions are only conditioned on past visits, that is embedding at time step $t$ can only attend to previous time steps $t-1, t-2 \dotsc$, we mask the attention layers with a causal triangular mask. This is the same "masked attention" in the decoder component of the original transformer network. This mask is applied to the set of embedding 

\begin{align}
&E = \{e_{c_1}, e_{c_2}, \dotsc e_{c_T}\}\\
&Attention(\bm{Q}, \bm{K}, \bm{V}) = \text{softmax}(\frac{\bm{Q}\bm{K}^T}{\sqrt{d}})\bm{V}
\end{align}

in the encoder block to ensure causality. Where the query $Q$, key $K$, and value $V$ are set to the sequence of embeddings $E$, and $d$ is the embedding dimension. To obtain the final code representation for timestep $t$, the $t^{th}$ output of the self-attention output is used. We call this code representation $E_{c_t}$.

\subsection{Medical Text Embedding}
%figure for text summarizer

\begin{figure}
\centering
\includegraphics[scale=1.8]{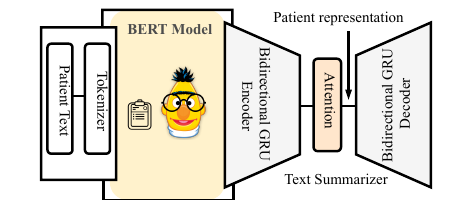}
\caption{The text representation module takes as input patient text and pre-processes them into tokens, which are embedded using BERT. Subsequently, it is fed to a text summarizer autoencoder network. In this work, a LSTM-AE is used to learn the intermediate patient representation.}
\label{fig:text_module}
\end{figure}

An overview of the text module is shown in Fig.~\ref{fig:text_module}. The embedding for the medical text sequence of a patient for time sequence $\{t_1, t_2, \dotsc, t_T\}$ is obtained by using a pre-trained BERT model initialized from BioBERT \cite{lee2019biobert} followed by a bidirectional GRU as a text summarizer. This is done, as the pre-trained BERT model has a fixed maximum sequence length of $n$ limiting the sequential scope of the text. Further, medical notes could get very lengthy during a visit and they contain different types of notes such as nurse notes, pharmacy notes, discharge notes, etc. the aggregate length of these at a time step $t$ could surpass the fixed length size of $n$. As the aim is to obtain a single visit representation, the aggregate notes at until time step $T$ are batched into a set of sentences $(u_1, u_2, \cdots, u_m)$, where $u_i \in \mathcal{Z}^n$ and $m$ is the maximum occurring length in the corpus after batching each visits text into sentences of $n$ words. The resulting set of sentences for a visit is embedded using the BERT model which results in a matrix $U \in \Re^{m \times d_{BERT}}$.
\begin{align}
    &h_{1:m} = GRU_{enc}(U_m, h_0)\\
    &E_{U_t} = \text{softmax}(\frac{\bm{h_{1:m}}\bm{h_{1:m}}^T}{\sqrt{d^{text}}})\bm{h_{1:m}}\\    
    &\hat{U} = GRU_{dec}(U_m, h_m)\\
    & L(U_{1:m, t}, \hat{U}_{1:m, t}) = \sum_i^m (u_i - \hat{u_i})^T(u_i - \hat{u_i}) 
\end{align}
Subsequently, the set of sentence representations are summarized into a single patient text representation using a text-summarizer module. This module follows an auto-encoder architecture with GRU's as the building block. The input is a set of sentence representations $\in \Re^{d_{text}}$ obtained by the BERT module applied on the aggregate text until time $t$ followed by a self-attention head on the hidden representations, where the decoder is tasked to output a sentence representations $\in \Re^{d_{text}}$ following the bottleneck. The objective of the summarizer is to reduce the MSE loss objective between the input sequence of the text embeddings and the models predicted representation at the correspondingly same time-step. Finally, the patients visit text representation is obtained by summarizing the set of sentence representations $\{U_1, U_2, ..., U_m\}$, where the output of the attention head applied on the encoders hidden representations is used as the representation.

\subsection{Patient Representation}

The final patient representation $Z^t$ at time $t$ is obtained by concatenating the code, and text representations. Additionally, the demographics $d_t$ of the patient recorded in visit at time $t$ is concatenated to the resulting vector. The demographics of patient contain information such as age, gender, race, etc, where categorical values are coded as one hot vectors. The final representation is denoted as $Z^t = [E_{c_t}; E_{U_t}; d_t]$, where the size of this vector is the sum of the components $d_{embedding} + d_{enc} + d_{demographics}$. This representation is used for downstream tasks.

We provide specific values for the dimensions of each component in our implementation details in the experiments section.

%Overview 

% figure transformer + bert + clasifier

%Code represenatation

%% unsupervised task

% Text representation

%% text summarization
\section{Experiments}
\subsection{Dataset}
We evaluate our model on the publicly available MIMIC-III clinical Database \cite{johnson2016mimic}. It consists of EHR records of 58,976 hospital admission consisting of 38,597 ICU patients from 2001 to 2012.  On average, each patient has 1.26 visits. The database contains tables associated with different data, where we extracted demographics, medical codes, and medical notes.
\subsubsection{Readmission task} The first task is to predict 30-day unplanned readmission to the ICU after being discharged. In this task, we formulate it as a binary problem, that is to predict whether a patient will be readmitted within 30-days after being discharged. Text entered into the MIMIC database, contains different reports, such as nurse notes, lab results, discharge summaries etc. We limit the text for each visit to contain the discharge summaries or text entered within the last 48h before the patient is discharged in the absence of discharge summaries.
\subsubsection{Mortality task} The second task is to predict mortality of patient, whether they passed away after being discharged or within the ICU. Similar to readmission it is formulated as a binary task. In this task, mortality related codes are discarded from the dataset and patients who were admitted for organ donations are removed. Additionally, the input text for each visit is limited to the first 24h of the admission.
\subsubsection{Length of stay task}: The third task is to forecast the length of stay (LOS) for patients. In this task, longer LOS is an indication of more severe illness and complex conditions. We formulate this problem as a multi-class classification problem by bucketing the length of stay into 9 classes: 1-7 correspond to one to seven days respectively, 8 corresponds to more than 1 week but less than 2, 9 corresponds to more than 2 weeks. The model is tasked to predict $P(Y=L|Z_t)$ where $L \in {1, 2, 3, 4, 5, 6, 7, 8, 9}$ denoting the different time intervals defined previously.
\subsubsection{Code prediction task}: In this task, clinical codes are predicted for new admissions of patients given past clinical codes and historical data of the patient. The predicted vector is high-dimensional equal to the size of unique codes.
\subsection{Data preprocessing} 
\begin{figure}[!tbp]
\centering
\includegraphics[scale=0.45]{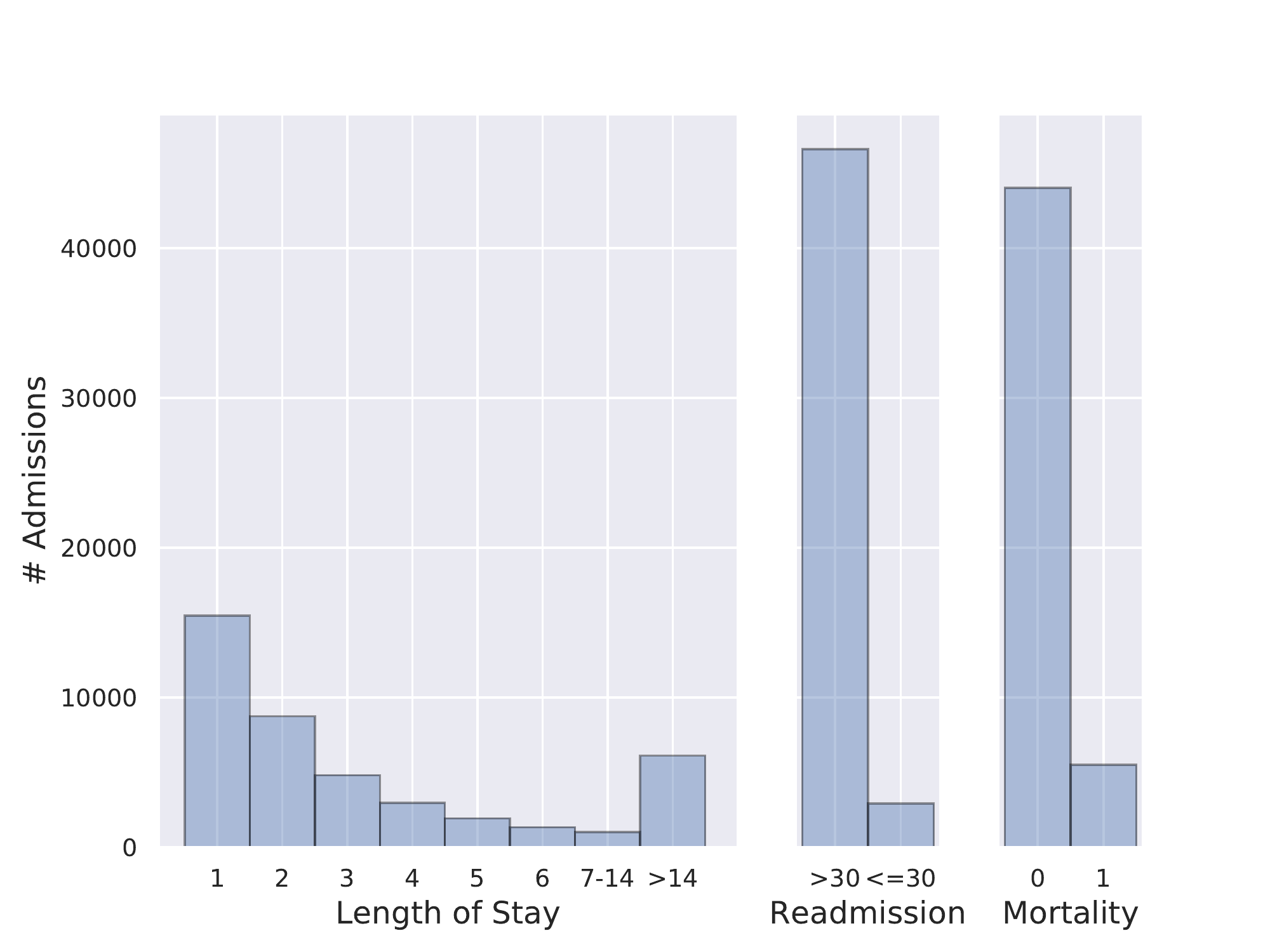}
\caption{Statistics of compiled cohort for both length of stay and readmission downstream tasks.}
\label{fig:stats}
\end{figure}

We extracted procedure, and diagnosis codes for each patient visit. These codes are defined by the International Classification of Disease (ICD9) and medications using the National Drug Code (NDC) standard. The total number of ICD9 codes in MIMIC-III is 6984, the number of drug codes is 3389, and the number of procedure codes is 1783. Codes whose frequency are less than $5$ are removed. We used the Clinical Classification Software for ICD9-CM\footnote{https://www.hcup-us.ahrq.gov/toolssoftware/ccs/ccs.jsp} to group the ICD9 diagnosis codes into 231 categories. The Clinical Classification Software for Services and Procedures\footnote{https://www.hcup-us.ahrq.gov/toolssoftware /ccs\_svcsproc/ccssvcproc.jsp} was used to group the procedure codes into 704 categories.  Additionally, patients of age under 18 were removed from the cohort. As medical notes contain many errors, we correct grammatical errors and remove non-alphanumerical characters. The text preprocessing closely follows \cite{clinicalbert}. After preprocessing, the average recorded number of codes per visit is 20.52,  the average number of words in medical notes is $7898$ and the average number of visits per patient is 1.29.
The statistics of the compiled cohort are depicted in Fig.\ref{fig:stats}

\subsection{Experimental Setup}
\subsubsection{Model Configuration \& training details} The training follows the method discussed in Section \ref{sec:method}. A medical concept model is trained independently on clinical codes in an unsupervised manner and similarly, the text summarizer is trained on the text portion. 

To train the transformer encoder we explored different values for hyperparameters, namely number of layers, number of multi-head attention heads $n_{head}$, dimension for each head $d_{head}$ and the final model representations $d_{code}$. We found $2$ layers perform well. We set the models representation dimension ($d_{code}$) to $128$. Further, the self-attention module contains 8 head ($n_{head}$), each with dimension 64 ($d_{head}$), which are a common configuration used in transformer networks. The network is trained using Adam \cite{kingma2014adam} with a cosine annealing schedule and with a period of 50 epochs. The initial learning rate is set to $0.00025$. The window size for the skipgram objective is set to $2$.

We initialize the BERT model with the pre-trained weights on medical nodes as presented in \cite{alsentzer2019publicly}. In this work, the BERT language model is initialized with BioBERT which is a model that has been trained on a large corpus of public medical data such as PubMed, medical abstracts, etc. Then the model was fine-tuned on the MIMIC-III clinical notes. 

To train the text summarizer we use a 2-layer bidirectional GRU autoencoder with the intermediate representation set to $d_{enc} = 128$. A teacher-forcing ratio of $0.5$ is used with a step learning rate schedule decay of $0.1$ every $50$ epochs with initial lr set to $10^{-3}$ \cite{lamb2016professor}. We have also tried using cosine annealing schedule, though this did not result in improvements.

Lastly, the classifier for downstream tasks is a simple 2-layer fully connected network. The first layer contains $\frac{d_{code} + d_{text} + d_{demographics}}{2}$ neurons with ReLU activation followed by a layer which maps to number of classes in the downstream task. When training on downstream tasks only the classifier weights are trained for 30 epochs with a step learning rate schedule decay of $0.1$ every $10$ epochs. This setting is used for all downstream tasks.

\subsubsection{Implementation details} We implemented all the models with Pytorch 1.0 \cite{paszke2017automatic}. For training the models we use the Adam optimizer \cite{kingma2014adam}.  In all experiments, the batch size is set to 32 on a machine equipped with 1 NIVIDIA 1080TI CUDA 9.0, 32GB Memory \& 8 CPU cores.

% Experiment 1 page
%% Dataset 
%% -- preprocessing
%% tasks
%% -- length of stay
%% -- readmission
%% -- recall @ 40 @30
%% Evaluation
%% -- roc curves
%% -- precision-recall auc
%% Experimental setup
\subsection{Evaluation Metrics} The compiled cohort consists of patientuids as keys that are unique to each patient which are used to create the test/train split. This is done using k-fold with $k=7$ on the patient keys, which the unsupervised models are trained on the train portion and validated on test ($\approx 15 \%$ of total data). The output for a particular time step is evaluated using the patient representation $Z_t$ at time $t$.
% For all experiments we use $15\%$ of the data selected randomly as test set. The rest are used for training. The presented values are the average of 5 experiments for each task.
\subsubsection{Area under the precision-recall (AU-PR)} this metric is the cumulative area under the curve by plotting precision and recall while varying the outputs $P(y_t=1|Z_t)$ true/false threshold from 0 to 1.
\subsubsection{Receiver operating characteristic curve (AU-ROC)} this metric is the area under the plot of the true positive rate against false positive rate while varying  outputs $P(y_t=1|Z_t)$ true/false threshold from 0 to 1.

\subsection{Baselines} We compare our model with the following baselines

% \subsubsection{Med2Vec \cite{choi2016multi}}
% A multi-layer perceptron is trained on medical codes using skip-gram model. An additional loss term is used for the co-occurrence of codes within the same visit as a regularization. The resulting output is a set of $\Re^d$ representation for each clinical code.

% \subsubsection{Time-aware embedding \cite{cai2018medical}}
% A multi-layer perceptron is trained on medical codes with an additional attention layer to take into account the temporal context of medical codes. The resulting model is trained using either skip-gram/CBOW. 

% \subsubsection{ClinicalBERT \cite{clinicalbert}}
% A BERT model is pre-trained on public medical data, which is then fine-tuned on clinical text. Following this pretraining a BERT classifier is initialized with pre-trained weights and further fine-tuned on downstream tasks. The input to this network is text.
\begin{itemize}
\item \textbf{Med2Vec} \cite{choi2016multi}: \\A multi-layer perceptron is trained on medical codes using the skip-gram objective function on a visit basis. An additional loss term is used for the co-occurrence of codes within the same visit as a regularization. The resulting output is a set of code representations in $\Re^d$.
\item \textbf{ClinicalBERT} \cite{clinicalbert}: \\A BERT model is pre-trained on public medical data, which is then fine-tuned on clinical text. Following this pre-training a BERT classifier is initialized with pre-trained weights and further fine-tuned on downstream tasks. The input to this network is text.
\item \textbf{Time-aware Embedding} \cite{cai2018medical}: \\A multi-layer perceptron is trained on medical codes with an additional attention layer to take into account the temporal context of medical codes. The resulting model is trained using either skip-gram/CBOW. 
\item \textbf{Patient2Vec} \cite{zhang2018patient2vec}: \\In this work a sequence of medical codes are embedded using word2vec model. The sequence of visits with irregular time intervals is then binned into a set of subsequences with standard intervals. Subsequently, the embedded vectors are stacked into a matrix where convolution stacked with GRU and attention models are applied to obtain the final patient representation.
\item \textbf{Joint-Skipgram}\cite{bai2018ehr}:\\
The embeddings are trained using both text and code as the vocabulary. In addition to the traditional skipgram loss, i.e. codes in the same visit predicting surrounding codes or text predicting surrounding text; the skipgram objective is modified such that text in a visit predict codes in the same visit and vice versa.
\item \textbf{Deepr}\cite{nguyen2016deepr}: \\
A set of clinical codes are embedded using skip-gram model. As visits contain multiple codes, the vectors corresponding to each code is stacked into a matrix, then the set of matrices for each visit is fed to a convolutional neural network and max-pooling layers to extract the final patient representation.
\item \textbf{$\bf Sg_{code} + Sg_{text}$}: \\
Embeddings for both code and text are learnt using the skipgram objective independently. Subsequently for downstream tasks a patient representation is obtained by concatenating the code and text embeddings.

\item \textbf{Supervised}: \\
The BERT model and summarizer takes as input raw text and transformer model raw codes, which is trained jointly on downstream tasks without pretraining.
\end{itemize}

We do not compare with more traditional text embeddings such bag of words (BOW) as other work have shown the benefits of using BERT as text representation in NLP tasks. 

We study the effect of different components presented by adding/removing text/code/demographics representation to our final patient visit representation.

\begin{table*}[!htb]
\caption{Code Recall}
\centering
\label{tab:code_recall_task}
\renewcommand{\arraystretch}{1.3}
\resizebox{2.0\columnwidth}{!}{

\begin{tabular}{l llllllll}
    \toprule
    \multicolumn{1}{c}{\bf Method} & \multicolumn{4}{c}{\bf Diagnosis@k } & \multicolumn{4}{c}{\bf Procedure@k } 
    \\ \hline 
    & 10 & 20 & 30 & 40 & 10 & 20 & 30 & 40\\ \hline \\
     \bf Ours & $47.92 \%_{\pm (0.6)}$&$65.11 \%_{\pm (0.4)}$&$75.25 \%_{\pm (0.3)}$&$82.00 \%_{\pm (0.4)}$ & $50.99 \%_{\pm (1.2)}$&$62.25 \%_{\pm (0.9)}$&$67.97 \%_{\pm (1.1)}$&$71.52 \%_{\pm (1.0)}$\\
     Joint-Skipgram \cite{bai2018ehr} & $41.16 \%_{\pm (0.6)}$&$59.49 \%_{\pm (0.7)}$&$71.26 \%_{\pm (0.7)}$&$79.73 \%_{\pm (0.5)}$& $50.88 \%_{\pm (0.9)}$&$61.33 \%_{\pm (0.7)}$&$67.32 \%_{\pm (0.9)}$&$70.95 \%_{\pm (0.7)}$\\
    Med2Vec \cite{choi2016multi} & $42.24 \%_{\pm (0.2)}$&$60.18 \%_{\pm (0.3)}$&$70.89 \%_{\pm (0.4)}$&$78.42 \%_{\pm (0.4)}$& $47.44 \%_{\pm (0.7)}$&$58.10 \%_{\pm (0.6)}$&$64.19 \%_{\pm (0.6)}$&$68.36 \%_{\pm (0.7)}$\\
    MCE \cite{cai2018medical} &$45.70 \%_{\pm (0.4)}$&$63.43 \%_{\pm (0.4)}$&$74.44 \%_{\pm (0.4)}$&$81.14 \%_{\pm (0.4)}$& $51.87 \%_{\pm (0.8)}$&$61.62 \%_{\pm (1.0)}$&$68.42 \%_{\pm (0.8)}$&$71.97 \%_{\pm (0.8)}$\\
    Deepr/P2V \cite{nguyen2016deepr, zhang2018patient2vec} & $35.30 \%_{\pm (0.3)}$&$52.40 \%_{\pm (0.4)}$&$65.04 \%_{\pm (0.4)}$&$74.14 \%_{\pm (0.3)}$& $42.47 \%_{\pm (0.7)}$&$55.59 \%_{\pm (1.1)}$&$62.42 \%_{\pm (0.8)}$&$66.49 \%_{\pm (0.9)}$\\
    Skipgram & $35.23 \%_{\pm (0.4)}$&$52.30 \%_{\pm (0.4)}$&$65.08 \%_{\pm (0.4)}$&$74.13 \%_{\pm (0.2)}$&$42.56 \%_{\pm (0.7)}$&$55.81 \%_{\pm (1.0)}$&$62.61 \%_{\pm (0.9)}$&$66.70 \%_{\pm (0.9)}$\\
    \bottomrule
\end{tabular}
}
\end{table*}

\section{Results}

% \begin{figure}[!tbp]
%   \centering
%   \begin{minipage}[b]{0.4\textwidth}
%     \includegraphics[width=\textwidth]{IEEEtran/drawio/los.pdf}
%   \end{minipage}
%   \hfill
%   \begin{minipage}[b]{0.4\textwidth}
%     \includegraphics[width=\textwidth]{IEEEtran/drawio/red.pdf}
%   \end{minipage}
%   \caption{Statistic of our dataset for both length of stay and readmission downstream tasks.}
% \end{figure}

% Results 1 page 
%% unsupervised tasks 
%% RED table
%% LOS table
To show the expressiveness of our representation, we evaluate its performance compared to baseline methods on downstream tasks and unsupervised learning tasks presented in the experiment section.  We present the results obtained by embedding both text and code as patient representation on three downstream tasks: (1) 30-day readmission, (2) mortality, and (3) length of stay (LOS). Note that in MIMIC-III database clinical codes are entered into the database upon discharge of a patient, as consequently the presented results may not be immediately clinically actionable in this case. Although, this may not be the case in other datasets where codes are updated throughout a visit. To this end the codes on the same visit are not fed to the model, but rather the previous set of codes are used.
% Further, we show interpretability of our method by using sensitivity analysis and plotting attention masks.

\subsection{Code pre-training} 

\begin{figure}
\centering
\includegraphics[scale=0.56]{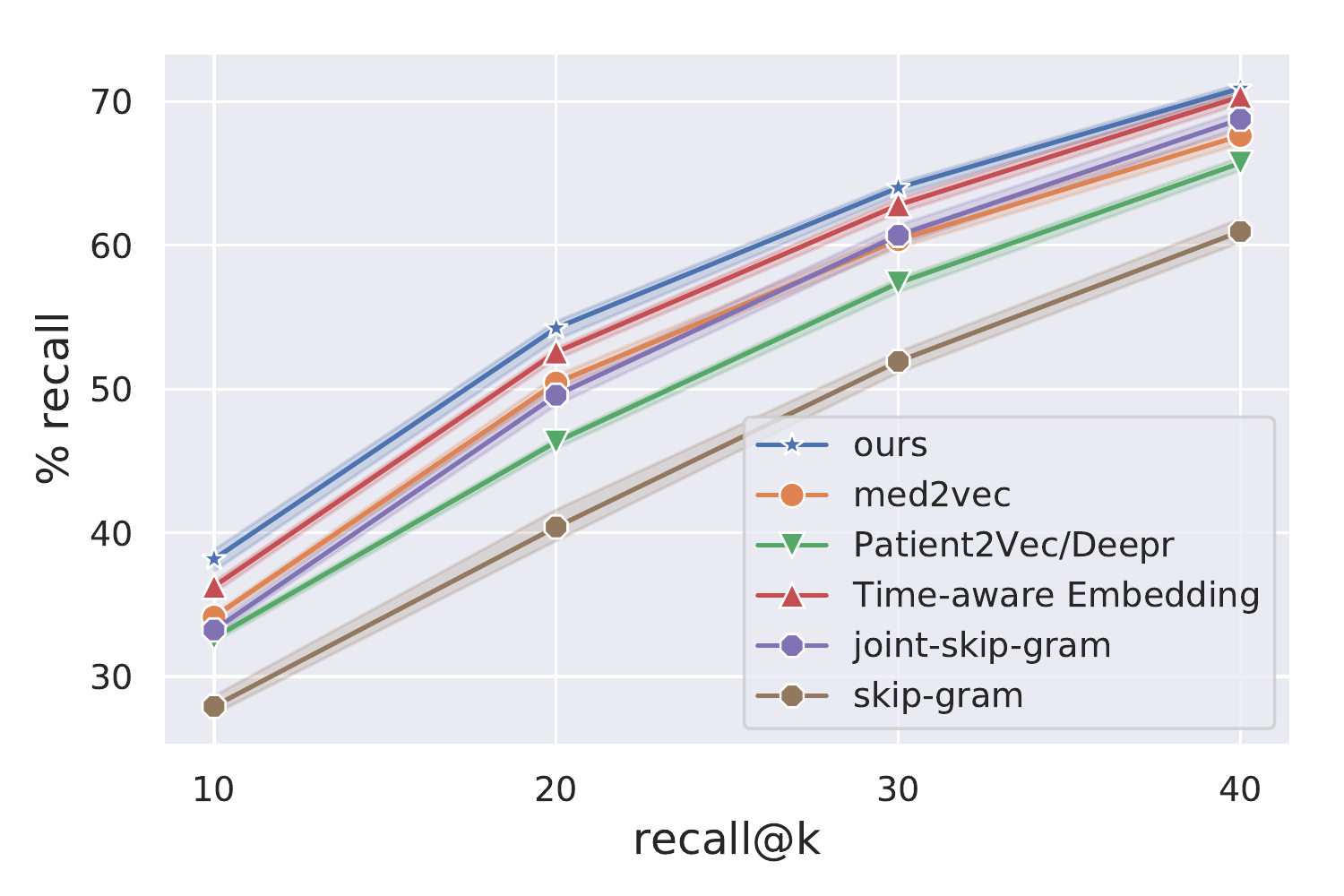}
\caption{Performance of different embedding schemes on next visit code prediction using recall @k.}
\label{fig:recall}
\end{figure}

The encoder network is trained on clinical concepts and as most patients have 3 hospital visits after filtering to atleast 2 visits the window size for the loss term is set to 2. The performance of our network is compared with baseline methods using recall@k. This metric is evaluated by computing 

$$\text{recall}@k = \frac{\text{\# of relevant codes in top k}}{\text{\# number of relevant code}}$$

This metric mimics a practitioners method of arriving at a diagnosis or prescribing medications where they generally have several sets of candidates as a presumed cause for the underlying condition of the patient. As shown in Fig. \ref{fig:recall} our code embedding consistently outperforms the baselines. All baselines are fine-tuned on the same corpora by exploring different architectural hyper-parameters, except the embedding size which is fixed to $d_{code} = 128$ for all models.
A comparison on recall for different codes (diagnosis, procedure) is also reported in Table. \ref{tab:code_recall_task}. From the results the time-aware code representations outperform other baselines.

\subsection{Ablation Study}

To evaluate the different components of the proposed method we conduct an ablation study on the inclusion/exclusion of the components in the final representation for downstream tasks. The complete representation using demographics, text representation and clinical code representation is the concatenation of these on a visit $Z^t = [E_{c_t}; E_{U_t}; d_t]$.

\subsubsection{Readmission} To better evaluate the effect of the different embedding components, we run an ablation study. The results are reported in table \ref{tab:readmission task}. 
≈
\begin{table}[!htb]
\caption{Downstream Tasks:  Readmission}
\centering
\label{tab:readmission task}
\renewcommand{\arraystretch}{1.3}

\begin{tabular}{l cccc}
    \toprule
    \multicolumn{1}{c}{\bf Method} &\multicolumn{1}{c}{\bf AUC-ROC } & \multicolumn{1}{c}{\bf PR-AUC } 
    \\ \hline \\
     \bf Text+Code+Demo & $\bf{{67.42\%}}$ & $\bf{68.03{\%}}$\\
     Text+Code & $65.74\%$ & $65.43\%$\\
    Text+Demo & $61.44\%$ & $62.81\%$\\
    Code+Demo & $64.53\%$ & $67.68\%$\\
    Text & $56.44\%$ & $56.55\%$\\
    Code & $60.74\%$ & $57.89\%$\\
    Demo & $54.76\%$ & $59.01\%$\\
    \bottomrule
\end{tabular}

\end{table}

From the results, it can be seen both text and code are informative for classifying readmission. The complete combination of text, code, demographics outperforms others in this case.\\

\subsubsection{Mortality}
Mortality task is concerned with predicting whether a patient will pass away within a pre-defined window. We predict mortality on a visit basis i.e. does the patient pass away in the current visit to the ICU. An ablation is done in table \ref{tab:mortality}. Similar, to previous ablations, the combination of text, code, and demographics outperforms other combinations.
\begin{table}[!htb]
\caption{Downstream Tasks:  Mortality}
\centering
\label{tab:mortality}

\renewcommand{\arraystretch}{1.3}
\begin{tabular}{l cccc}
    \toprule
    \multicolumn{1}{c}{\bf Model} &\multicolumn{1}{c}{\bf AUC-ROC } & \multicolumn{1}{c}{\bf PR-AUC } 
    \\ \hline \\
     \bf Text+Code+Demo & $\bf{63.42\%}$ & $\bf65.65{\%}$\\
    Text+Code & $59.75\%$ & ${60.33\%}$\\
    Text+Demo & $61.54\%$ & $64.07\%$\\
    Code+Demo & $60.41\%$ & $64.41\%$\\
    Text & $57.14\%$ & $57.72\%$\\
    Code & $56.91\%$ & $53.71\%$\\
    Demo & $60.10\%$ & $62.02\%$\\
    \bottomrule
\end{tabular}

\end{table}
\subsubsection{Length of Stay}
In general length of stay is a much more challenging task compared to readmission binary task. In this classification task, we limit the medical text to the first 24-hours of the current patient visit in which length of stay is being predicted. Limiting the note context window is done as medical text could include information on date patient has been discharged. In this task the network is trained on imbalanced class data split and tested on balanced data, this is done as the majority of classes are discharged within 24h-48h.
\begin{table}[!htb]
\caption{Downstream Tasks: Length of Stay Top-1 avg over 5-fold cross validation}
\centering

\label{tab:ablation_los}
\renewcommand{\arraystretch}{1.3}

\begin{tabular}{l cccc}
    \toprule
    \multicolumn{1}{c}{\bf Method} &\multicolumn{1}{c}{\bf Top-1} 
    \\ \hline \\
    \bf Text+Code+Demo & $\bf{25.57\%}$ \\
    Text+Code & $23.38\%$ \\
    Text+Demo & $21.10\%$ \\
    Code+Demo & $22.22\%$  \\
    Text & $18.82\%$  \\
    Code & $20.54\%$ \\
    Demo & $20.13\%$ \\
    \bottomrule
\end{tabular}

\end{table}

% \begin{figure}[htb]
%     \centering
%     \includegraphics[scale=0.5]{IEEEtran/drawio/confusionmatrix.pdf}
%     \caption{Normalized confusion matrix for the length of stay downstream task.}
%     \label{fig:confusion}
% \end{figure}
% We plot the confusion matrix for the best performing classifier using text, code, and demographics in Fig. \ref{fig:confusion}. The classifier is predicting at extremities of either 1-2 day length of stay or more than 2 weeks.

As shown in table \ref{tab:ablation_los}, the combination of text, code and demo outperforms others. We conclude in this set of experiments, solely using text and diagnosis codes is not predictive enough to forcast the length of stay of patients.

\subsubsection{Comparison with Other Work}

\begin{table*}[h]
\caption{Comparison with other work on downstream tasks.}
\label{tab:comparison}
\begin{center}
\renewcommand{\arraystretch}{1.3}
% \resizebox{\columnwidth}{!}{%

\begin{tabular}{l ccccc}
\toprule
 \multicolumn{1}{c}{\bf Method} &\multicolumn{2}{c}{\bf Readmisison} &\multicolumn{2}{c}{\bf Mortality} &\multicolumn{1}{c}{\bf LOS}
\\ \hline 
& ROC & PR-AUC & ROC & PR-AUC & ACC \\\hline\\ 
\bf Ours & $67.42\% {\footnotesize \pm (2.1)}$ & $68.03\% {\footnotesize \pm (1.5)}$ & $63.42\% {\footnotesize \pm (0.6)}$ & $65.65\% {\footnotesize \pm (0.9)}$  & $25.57\% {\footnotesize \pm (2.7)}$\\
ClinicalBert \cite{clinicalbert} & $64.41\% {\footnotesize \pm (2.3)}$ & $67.71\% {\footnotesize \pm (1.3)}$ & $61.23\% {\footnotesize \pm (1.9)}$ & $64.83\% {\footnotesize \pm (1.1)}$ & $22.65\% {\footnotesize \pm (1.6)}$ \\
Joint-Skipgram$^*$ \cite{bai2018ehr} & $64.59\% {\footnotesize \pm (2.9)}$ & $65.27\% {\footnotesize \pm (2.7)}$ & $61.73\% {\footnotesize \pm (1.6)}$ & $65.03\% {\footnotesize \pm (1.1)}$ & $23.27\% {\footnotesize \pm (1.6)}$ \\
SG$_{code}$ + SG$_{text}$$^*$ & $65.79\% {\footnotesize \pm (1.9)}$ & $66.72\% {\footnotesize \pm (0.6)}$ & $62.23\% {\footnotesize \pm (0.6)}$ & $64.77\% {\footnotesize \pm (1.1)}$ & $24.54\% {\footnotesize \pm (0.4)}$ \\
MCE$^*$ \cite{cai2018medical} & $64.61\% {\footnotesize \pm (2.1)}$ & $63.96\% {\footnotesize \pm (2.1)}$ & $62.72\% {\footnotesize \pm (2.1)}$ & $ 64.78\% {\footnotesize \pm (2.2)}$  & $22.47 \% {\footnotesize \pm (2.3)}$\\
Deepr$^*$ \cite{nguyen2016deepr} & $62.63\% {\footnotesize \pm (1.9)}$ & $62.37\% {\footnotesize \pm (1.5)}$ & $60.49\% {\footnotesize \pm (1.2)}$ & $61.45\% {\footnotesize \pm (1.5)}$  & $22.81\% {\footnotesize \pm (0.5)}$ \\
Patient2Vec $^*$ \cite{zhang2018patient2vec} & $56.13\% {\footnotesize \pm (3.3)}$ & $65.12\% {\footnotesize \pm (1.1)}$ & $61.19\% {\footnotesize \pm (1.4)}$ & $62.51\% {\footnotesize \pm (1.6)}$  & $21.46\% {\footnotesize \pm (1.2)}$\\
Med2Vec \cite{choi2016multi} & $63.44\% {\footnotesize \pm (2.8)}$ & $62.77\% {\footnotesize \pm (2.2)}$ & $61.67\% {\footnotesize \pm (1.6)}$ & $63.22\% {\footnotesize \pm (0.8)}$  & $24.15\% {\footnotesize \pm (0.4)}$\\
Ours - Supervised & $65.38\% {\footnotesize \pm (3.4)}$ & $66.07\% {\footnotesize \pm (2.4)}$ & $62.26\% {\footnotesize \pm (1.7)}$ & $64.32\% {\footnotesize \pm (2.3)}$  & $25.12\% {\footnotesize \pm (3.2)}$\\
\bottomrule
\end{tabular}%
% }

{1. Methods marked * is our best-effort re-implementation due to unavailability of source. \par}

\end{center}
\end{table*}

We ran all baseline models on three downstream tasks using the same pre-processing steps. The final results are reported in table \ref{tab:comparison}. To compare our method, we use the combination of text, code, and demographics. As demographics has proven to have predictive power for the downstream tasks, other methods are augmented to use this as input/concatenated to the output representation prior to prediction.  From Table. \ref{tab:comparison} the presented method outperforms others by a 1-2\% margin, on all tasks demonstrating the usefulness of unsupervised pre-training.

%\FloatBarrier
\section{Conclusion}

Effective representation learning for EHR data is an essential step to improving care. We study embedding both medical codes and notes into a unified vector representation for downstream task prediction. The presented method effectively takes the temporal context of these two data streams and provides a patient visit representation. The proposed method was evaluated on three tasks namely, readmission, mortality, and length of stay outperforming other methods. An ablation study was also done showing the usefulness of both text and code when modeling patient visits. Future work could focus on adding additional data streams to the pipeline by taking into account real-time vitals and measurements taken from a patient as they undergo care. 

\bibliographystyle{IEEEtran}
% argument is your BibTeX string definitions and bibliography database(s)
\bibliography{bib}

% Generated by IEEEtran.bst, version: 1.14 (2015/08/26)
\begin{thebibliography}{10}
\providecommand{\url}[1]{#1}
\csname url@samestyle\endcsname
\providecommand{\newblock}{\relax}
\providecommand{\bibinfo}[2]{#2}
\providecommand{\BIBentrySTDinterwordspacing}{\spaceskip=0pt\relax}
\providecommand{\BIBentryALTinterwordstretchfactor}{4}
\providecommand{\BIBentryALTinterwordspacing}{\spaceskip=\fontdimen2\font plus
\BIBentryALTinterwordstretchfactor\fontdimen3\font minus
  \fontdimen4\font\relax}
\providecommand{\BIBforeignlanguage}[2]{{%
\expandafter\ifx\csname l@#1\endcsname\relax
\typeout{** WARNING: IEEEtran.bst: No hyphenation pattern has been}%
\typeout{** loaded for the language `#1'. Using the pattern for}%
\typeout{** the default language instead.}%
\else
\language=\csname l@#1\endcsname
\fi
#2}}
\providecommand{\BIBdecl}{\relax}
\BIBdecl

\bibitem{teres1987validation}
D.~Teres, S.~Lemeshow, J.~S. Avrunin, and H.~Pastides, ``Validation of the
  mortality prediction model for icu patients.'' \emph{Critical care medicine},
  vol.~15, no.~3, pp. 208--213, 1987.

\bibitem{campbell2008predicting}
A.~J. Campbell, J.~A. Cook, G.~Adey, and B.~H. Cuthbertson, ``Predicting death
  and readmission after intensive care discharge,'' \emph{British journal of
  anaesthesia}, vol. 100, no.~5, pp. 656--662, 2008.

\bibitem{bengio2013representation}
Y.~Bengio, A.~Courville, and P.~Vincent, ``Representation learning: A review
  and new perspectives,'' \emph{IEEE transactions on pattern analysis and
  machine intelligence}, vol.~35, no.~8, pp. 1798--1828, 2013.

\bibitem{mikolov2013}
\BIBentryALTinterwordspacing
T.~Mikolov, I.~Sutskever, K.~Chen, G.~S. Corrado, and J.~Dean, ``Distributed
  representations of words and phrases and their compositionality,'' in
  \emph{Advances in Neural Information Processing Systems 26}, C.~J.~C. Burges,
  L.~Bottou, M.~Welling, Z.~Ghahramani, and K.~Q. Weinberger, Eds.\hskip 1em
  plus 0.5em minus 0.4em\relax Curran Associates, Inc., 2013, pp. 3111--3119.
  [Online]. Available:
  \url{http://papers.nips.cc/paper/5021-distributed-representations-of-words-and-phrases-and-their-compositionality.pdf}
\BIBentrySTDinterwordspacing

\bibitem{DBLP:journals/corr/abs-1806-02873}
\BIBentryALTinterwordspacing
X.~Cai, J.~Gao, K.~Y. Ngiam, B.~C. Ooi, Y.~Zhang, and X.~Yuan, ``Medical
  concept embedding with time-aware attention,'' \emph{CoRR}, vol.
  abs/1806.02873, 2018. [Online]. Available:
  \url{http://arxiv.org/abs/1806.02873}
\BIBentrySTDinterwordspacing

\bibitem{cai2018medical}
------, ``Medical concept embedding with time-aware attention,'' \emph{arXiv
  preprint arXiv:1806.02873}, 2018.

\bibitem{bahdanau2014neural}
D.~Bahdanau, K.~Cho, and Y.~Bengio, ``Neural machine translation by jointly
  learning to align and translate,'' \emph{arXiv preprint arXiv:1409.0473},
  2014.

\bibitem{luong2015effective}
M.-T. Luong, H.~Pham, and C.~D. Manning, ``Effective approaches to
  attention-based neural machine translation,'' \emph{arXiv preprint
  arXiv:1508.04025}, 2015.

\bibitem{vaswani2017attention}
A.~Vaswani, N.~Shazeer, N.~Parmar, J.~Uszkoreit, L.~Jones, A.~N. Gomez,
  {\L}.~Kaiser, and I.~Polosukhin, ``Attention is all you need,'' in
  \emph{Advances in neural information processing systems}, 2017, pp.
  5998--6008.

\bibitem{rumelhart1988learning}
D.~E. Rumelhart, G.~E. Hinton, R.~J. Williams \emph{et~al.}, ``Learning
  representations by back-propagating errors,'' \emph{Cognitive modeling},
  vol.~5, no.~3, p.~1, 1988.

\bibitem{bengio2003neural}
Y.~Bengio, R.~Ducharme, P.~Vincent, and C.~Jauvin, ``A neural probabilistic
  language model,'' \emph{Journal of machine learning research}, vol.~3, no.
  Feb, pp. 1137--1155, 2003.

\bibitem{choi2016multi}
E.~Choi, M.~T. Bahadori, E.~Searles, C.~Coffey, M.~Thompson, J.~Bost,
  J.~Tejedor-Sojo, and J.~Sun, ``Multi-layer representation learning for
  medical concepts,'' in \emph{Proceedings of the 22nd ACM SIGKDD International
  Conference on Knowledge Discovery and Data Mining}.\hskip 1em plus 0.5em
  minus 0.4em\relax ACM, 2016, pp. 1495--1504.

\bibitem{nguyen2016deepr}
P.~Nguyen, T.~Tran, N.~Wickramasinghe, and S.~Venkatesh, ``$\mathtt {Deepr}$: a
  convolutional net for medical records,'' \emph{IEEE journal of biomedical and
  health informatics}, vol.~21, no.~1, pp. 22--30, 2016.

\bibitem{Choi:2017:GGA:3097983.3098126}
\BIBentryALTinterwordspacing
E.~Choi, M.~T. Bahadori, L.~Song, W.~F. Stewart, and J.~Sun, ``Gram:
  Graph-based attention model for healthcare representation learning,'' in
  \emph{Proceedings of the 23rd ACM SIGKDD International Conference on
  Knowledge Discovery and Data Mining}, ser. KDD '17.\hskip 1em plus 0.5em
  minus 0.4em\relax New York, NY, USA: ACM, 2017, pp. 787--795. [Online].
  Available: \url{http://doi.acm.org/10.1145/3097983.3098126}
\BIBentrySTDinterwordspacing

\bibitem{zhang2018patient2vec}
J.~Zhang, K.~Kowsari, J.~H. Harrison, J.~M. Lobo, and L.~E. Barnes,
  ``Patient2vec: A personalized interpretable deep representation of the
  longitudinal electronic health record,'' \emph{IEEE Access}, vol.~6, pp.
  65\,333--65\,346, 2018.

\bibitem{MIME2018}
\BIBentryALTinterwordspacing
E.~Choi, C.~Xiao, W.~Stewart, and J.~Sun, ``Mime: Multilevel medical embedding
  of electronic health records for predictive healthcare,'' in \emph{Advances
  in Neural Information Processing Systems 31}, S.~Bengio, H.~Wallach,
  H.~Larochelle, K.~Grauman, N.~Cesa-Bianchi, and R.~Garnett, Eds.\hskip 1em
  plus 0.5em minus 0.4em\relax Curran Associates, Inc., 2018, pp. 4547--4557.
  [Online]. Available:
  \url{http://papers.nips.cc/paper/7706-mime-multilevel-medical-embedding-of-electronic-health-records-for-predictive-healthcare.pdf}
\BIBentrySTDinterwordspacing

\bibitem{galvan-etal-2018-investigating}
\BIBentryALTinterwordspacing
D.~Galvan, N.~Okazaki, K.~Matsuda, and K.~Inui, ``Investigating the challenges
  of temporal relation extraction from clinical text,'' in \emph{Proceedings of
  the Ninth International Workshop on Health Text Mining and Information
  Analysis}.\hskip 1em plus 0.5em minus 0.4em\relax Brussels, Belgium:
  Association for Computational Linguistics, Oct. 2018, pp. 55--64. [Online].
  Available: \url{https://www.aclweb.org/anthology/W18-5607}
\BIBentrySTDinterwordspacing

\bibitem{textrepresentation}
\BIBentryALTinterwordspacing
D.~Dligach and T.~A. Miller, ``Learning patient representations from text,''
  \emph{CoRR}, vol. abs/1805.02096, 2018. [Online]. Available:
  \url{http://arxiv.org/abs/1805.02096}
\BIBentrySTDinterwordspacing

\bibitem{liu2018deep}
J.~Liu, Z.~Zhang, and N.~Razavian, ``Deep ehr: Chronic disease prediction using
  medical notes,'' \emph{arXiv preprint arXiv:1808.04928}, 2018.

\bibitem{devlin2018bert}
J.~Devlin, M.-W. Chang, K.~Lee, and K.~Toutanova, ``Bert: Pre-training of deep
  bidirectional transformers for language understanding,'' \emph{arXiv preprint
  arXiv:1810.04805}, 2018.

\bibitem{clinicalbert}
K.~Huang, J.~Altosaar, and R.~Ranganath, ``Clinicalbert: Modeling clinical
  notes and predicting hospital readmission,'' \emph{arXiv:1904.05342}, 2019.

\bibitem{bai2018ehr}
T.~Bai, A.~K. Chanda, B.~L. Egleston, and S.~Vucetic, ``Ehr phenotyping via
  jointly embedding medical concepts and words into a unified vector space,''
  \emph{BMC medical informatics and decision making}, vol.~18, no.~4, p. 123,
  2018.

\bibitem{mullenbach2018explainable}
J.~Mullenbach, S.~Wiegreffe, J.~Duke, J.~Sun, and J.~Eisenstein, ``Explainable
  prediction of medical codes from clinical text,'' \emph{arXiv preprint
  arXiv:1802.05695}, 2018.

\bibitem{DBLP:journals/corr/abs-1810-04805}
\BIBentryALTinterwordspacing
J.~Devlin, M.~Chang, K.~Lee, and K.~Toutanova, ``{BERT:} pre-training of deep
  bidirectional transformers for language understanding,'' \emph{CoRR}, vol.
  abs/1810.04805, 2018. [Online]. Available:
  \url{http://arxiv.org/abs/1810.04805}
\BIBentrySTDinterwordspacing

\bibitem{vincent2008extracting}
P.~Vincent, H.~Larochelle, Y.~Bengio, and P.-A. Manzagol, ``Extracting and
  composing robust features with denoising autoencoders,'' in \emph{Proceedings
  of the 25th international conference on Machine learning}.\hskip 1em plus
  0.5em minus 0.4em\relax ACM, 2008, pp. 1096--1103.

\bibitem{mkdenoising}
\BIBentryALTinterwordspacing
M.~Kachuee, S.~Darabi, B.~Moatamed, and M.~Sarrafzadeh, ``Dynamic feature
  acquisition using denoising autoencoders,'' \emph{CoRR}, vol. abs/1811.01249,
  2018. [Online]. Available: \url{http://arxiv.org/abs/1811.01249}
\BIBentrySTDinterwordspacing

\bibitem{lee2019biobert}
J.~Lee, W.~Yoon, S.~Kim, D.~Kim, S.~Kim, C.~H. So, and J.~Kang, ``Biobert:
  pre-trained biomedical language representation model for biomedical text
  mining,'' \emph{arXiv preprint arXiv:1901.08746}, 2019.

\bibitem{johnson2016mimic}
A.~E. Johnson, T.~J. Pollard, L.~Shen, H.~L. Li-wei, M.~Feng, M.~Ghassemi,
  B.~Moody, P.~Szolovits, L.~A. Celi, and R.~G. Mark, ``Mimic-iii, a freely
  accessible critical care database,'' \emph{Scientific data}, vol.~3, p.
  160035, 2016.

\bibitem{kingma2014adam}
D.~P. Kingma and J.~Ba, ``Adam: A method for stochastic optimization,''
  \emph{arXiv preprint arXiv:1412.6980}, 2014.

\bibitem{alsentzer2019publicly}
E.~Alsentzer, J.~R. Murphy, W.~Boag, W.-H. Weng, D.~Jin, T.~Naumann, and
  M.~McDermott, ``Publicly available clinical bert embeddings,'' \emph{arXiv
  preprint arXiv:1904.03323}, 2019.

\bibitem{lamb2016professor}
A.~M. Lamb, A.~G. A.~P. Goyal, Y.~Zhang, S.~Zhang, A.~C. Courville, and
  Y.~Bengio, ``Professor forcing: A new algorithm for training recurrent
  networks,'' in \emph{Advances In Neural Information Processing Systems},
  2016, pp. 4601--4609.

\bibitem{paszke2017automatic}
A.~Paszke, S.~Gross, S.~Chintala, G.~Chanan, E.~Yang, Z.~DeVito, Z.~Lin,
  A.~Desmaison, L.~Antiga, and A.~Lerer, ``Automatic differentiation in
  {PyTorch},'' in \emph{NIPS Autodiff Workshop}, 2017.

\end{thebibliography}
%
% <OR> manually copy in the resultant .bbl file
% set second argument of \begin to the number of references
% (used to reserve space for the reference number labels box)

% biography section
% 
% If you have an EPS/PDF photo (graphicx package needed) extra braces are
% needed around the contents of the optional argument to biography to prevent
% the LaTeX parser from getting confused when it sees the complicated
% \includegraphics command within an optional argument. (You could create
% your own custom macro containing the \includegraphics command to make things
% simpler here.)
%\begin{IEEEbiography}[{\includegraphics[width=1in,height=1.25in,clip,keepaspectratio]{mshell}}]{Michael Shell}
% or if you just want to reserve a space for a photo:

% \begin{IEEEbiography}{Michael Shell}
% Biography text here.
% \end{IEEEbiography}

% % if you will not have a photo at all:
% \begin{IEEEbiographynophoto}{John Doe}
% Biography text here.
% \end{IEEEbiographynophoto}

% % insert where needed to balance the two columns on the last page with
% % biographies
% %\newpage

% \begin{IEEEbiographynophoto}{Jane Doe}
% Biography text here.
% \end{IEEEbiographynophoto}

% You can push biographies down or up by placing
% a \vfill before or after them. The appropriate
% use of \vfill depends on what kind of text is
% on the last page and whether or not the columns
% are being equalized.

%\vfill

% Can be used to pull up biographies so that the bottom of the last one
% is flush with the other column.
%\enlargethispage{-5in}

% that's all folks
\end{document}